\documentclass{llncs} %

\usepackage{float}
\usepackage{graphicx}
\usepackage{wrapfig}

\usepackage[labelfont=bf]{caption}
\usepackage{amsmath} 
\usepackage{mathtools} 
\usepackage{amsfonts} 
\usepackage{amssymb}
\usepackage{comment}

\usepackage{array}
\usepackage{booktabs}
\usepackage{rotating}
\usepackage{multirow}
\usepackage{multicol}
\usepackage{lscape}
\usepackage{subfig}

\usepackage[svgnames]{xcolor}

\definecolor{babyblue}{rgb}{0.54, 0.81, 0.94}
\definecolor{armygreen}{rgb}{0.29, 0.33, 0.13}
\definecolor{brightlavender}{rgb}{0.75, 0.58, 0.89}
\definecolor{aqua}{rgb}{0.0, 1.0, 1.0}
\definecolor{caribbeangreen}{rgb}{0.0, 0.8, 0.6}
\definecolor{reddish}{rgb}{0.82, 0.1, 0.26}
\definecolor{emerald}{rgb}{0.31, 0.78, 0.47}
\definecolor{jasper}{rgb}{0.84, 0.23, 0.24}
\definecolor{red}{rgb}{1.0, 0.0, 0.0}
\definecolor{green}{rgb}{0.0, 1.0, 0.0}
\definecolor{blue}{rgb}{0.0, 0.0, 1.0}
\definecolor{darkgreen}{rgb}{0.1, 0.7, 0.1}
\definecolor{darkblue}{rgb}{0.1, 0.1, 0.7}
\definecolor{red}{rgb}{0.7, 0.1, 0.1}
	
\usepackage[export]{adjustbox}
\usepackage[colorlinks=true,linkcolor=caribbeangreen,citecolor=emerald]{hyperref}
\usepackage[colorinlistoftodos,prependcaption,textsize=tiny]{todonotes}

\usepackage{algorithm,algorithmicx,algpseudocode}

\usepackage{color,soul}

\usepackage{tikz,xcolor,hyperref}

\definecolor{lime}{HTML}{A6CE39}
\DeclareRobustCommand{\orcidicon}{
	\begin{tikzpicture}
	\draw[lime, fill=lime] (0,0) 
	circle [radius=0.16] 
	node[white] {{\fontfamily{qag}\selectfont \tiny ID}};
	\draw[white, fill=white] (-0.0625,0.095) 
	circle [radius=0.007];
	\end{tikzpicture}
	\hspace{-2mm}
}

\foreach \x in {A, ..., Z}{\expandafter\xdef\csname orcid\x\endcsname{\noexpand\href{https://orcid.org/\csname orcidauthor\x\endcsname}
			{\noexpand\orcidicon}}
}
			
\begin{document}

\title{Multi-Scale Profiling of Brain Multigraphs by Eigen-based Cross-Diffusion and Heat Tracing for Brain State Profiling}

\titlerunning{Short Title}  

\author{Mustafa Sa\u{g}lam \index{Sa\u{g}lam, Mustafa}  \and Islem Rekik\orcidA{} \index{Rekik, Islem}\thanks{ {corresponding author: irekik@itu.edu.tr, \url{http://basira-lab.com}, GitHub: \url{http://github.com/basiralab}. } This work is accepted for publication in the GRaphs in biomedicAl Image anaLysis (GRAIL) workshop Springer proceedings in conjunction with MICCAI 2020.}}

\institute{BASIRA Lab, Faculty of Computer and Informatics, Istanbul Technical University, Istanbul, Turkey}

\authorrunning{M. Sa\u{g}lam et al.}

\maketitle              

\begin{abstract}

The individual brain can be viewed as a highly-complex \emph{multigraph} (i.e. a set of graphs also called connectomes), where each graph represents a unique connectional view of pairwise brain region (node) relationships such as function or morphology. Due to its multi-fold complexity, understanding how brain disorders alter not only a single view of the brain graph, but its \emph{multigraph representation} at the individual and population scales, remains one of the most challenging obstacles to profiling brain connectivity for ultimately disentangling a wide spectrum of brain states (e.g., healthy vs. disordered). Existing graph theory based works on comparing brain graphs in different states have major drawbacks. \emph{First}, these techniques are conventionally designed to operate on single brain graphs, while brain multigraph representations remain widely untapped. \emph{Second}, the bulk of such works lies in using graph comparison techniques such as kernel-based or graph distance editing methods, which fail to simultaneously satisfy graph scalability, node- and permutation-invariance criteria. To address these limitations and while cross-pollinating the fields of spectral graph theory and diffusion models, we unprecedentedly propose an eigen-based cross-diffusion strategy for multigraph brain integration, comparison, and profiling. Specifically, we first devise a brain multigraph fusion model guided by eigenvector centrality to rely on most central nodes in the cross-diffusion process. Next, since the graph spectrum encodes its shape (or geometry) as if one can hear the shape of the graph, for the first time, we profile the \emph{fused multigraphs} at several diffusion timescales by extracting the compact heat-trace signatures of their corresponding Laplacian matrices. Such brain multigraph heat-trace  profiles nicely satisfy the three graph comparison criteria. More importantly, we reveal for the first time autistic and healthy profiles of morphological brain multigraphs, derived from T1-w magnetic resonance imaging (MRI), and demonstrate their discriminability in boosting the classification of unseen samples in comparison with state-of-the-art methods. This study presents the first step towards hearing the shape of the brain multigraph that can be leveraged for profiling and disentangling comorbid neurological disorders, thereby advancing precision medicine.

\end{abstract}


\section{Introduction}

The development of network neuroscience \cite{Bassett:2017} aims to present a holistic picture of the brain graph (also called network or connectome), a universal representation of heterogeneous pairwise brain region relationships (e.g., correlation in neural activity or dissimilarity in morphology). Due to its multi-fold complexity, the underlying causes of neurological and psychiatric disorders, such as Alzheimer's disease, autism, and depression remain largely unknown and difficult to pin down \cite{Fornito:2015,Heuvel:2019}. How these brain disorders unfold at the individual and population scales remains one of the most challenging obstacles to understanding how the \emph{brain graph} gets altered by disorders, let alone a \emph{brain multigraph}. Indeed, using different measurements, one can build a brain multigraph, composed of a set of graphs, each capturing a unique view of the brain construct (such as morphology or function) \cite{Bassett:2017,Corps:2019,Bilgen:2020}. Profiling brain multigraphs remains a formidable challenge to identify the most representative and shared brain alterations caused by a specific disorder, namely \emph{`disorder profile'}, in a population of brain multi-graphs. Such integral profile can be revealed by what we name as \emph{multigraph brain profile}, which would constitute an unprecedented contribution to network neuroscience and brain mapping literature as it would chart the connectional geography of the brain. 

Estimating such profiles highly depends on using reliable graph comparison techniques. However, existing graph theory based works on comparing brain graphs in different states have major drawbacks. \emph{First}, these techniques are conventionally designed to operate on single brain graphs, while brain multigraph representations remain widely untapped. \emph{Second}, the bulk of such works lies in using graph comparison techniques such as kernel-based or graph distance editing methods, which fail to simultaneously satisfy graph scalability, node- and permutation-invariance criteria. For instance, one can use graph edit distance (GED) technique \cite{Sanfeliu:1983} that estimates the minimal number of edit operations needed to transform a graph into another. However, this is an NP hard problem that becomes intractable when scaling up graph sizes. Graph multiple kernel-based comparison methods, on the other hand, are more natural when desiring scale-adaptivity since each kernel can capture a particular graph scale such as the multi-scale Laplacian graph kernel method proposed in \cite{Kondor:2016}. However, such techniques raise a computational overhead cubic in deriving Laplacian matrix eigenvalues and when the size of the graph exponentially grows. Traditional statistical methods including the family of spectral distances (FGSD) \cite{Verma:2017} produces a high-dimensional sparse representation as a histogram on the dense biharmonic graph kernel; however, such methods are not scale-adaptive and are also inapplicable to reasonably large graphs due to their quadratic time complexity. 

Adding to the difficulty of profiling the state of a single brain graph, profiling a population of brain \emph{multigraphs}, to eventually discover disorder-specific profiles, presents a big jump in the field of network neuroscience, which we set out to take in this paper. Specifically, while addressing the aforementioned limitations and while cross-pollinating the fields of spectral graph theory and diffusion models, we unprecedentedly propose an eigen-based cross-diffusion strategy for brain multigraph \emph{integration, comparison, and profiling}. In the first step, we aim to learn how to fuse a population of brain multigraphs into a single graph by capitalizing on unsupervised graph diffusion and fusion technique presented in  \cite{Wang:2012}. However, while cross-diffusing a set of graphs for eventually estimating a representative integral graph representation of each individual brain multigraph, \cite{Wang:2012} overlooks the topological properties of graph nodes such as node centrality, which better capture local and global structure of the brain connectivity providing a more holistic measurement of the brain graph. To address this limitation, we propose a novel multigraph cross-diffusion based on a graph Laplacian derived from eigen-centrality measures. In the second step, since a graph spectrum encodes its shape (or geometry) as if one can hear the shape of the graph \cite{Kac:1966}, for the first time, we profile the \emph{fused multigraphs} at several diffusion timescales by extracting the compact heat-trace signatures of their corresponding Laplacian matrices. To this aim, we adopt network Laplacian spectral descriptor (NetLSD) introduced in \cite{Tsitsulin:2018} to produce brain multigraph heat-trace profiles,  which nicely satisfy permutation- and size-invariance, and scale-adaptivity. As one can ``hear'' the connectivity of the drum  if we were to represent its shape as a graph \cite{Kac:1966}, in this paper, we hear the connectivity of autistic and healthy morphological brain multigraphs, derived from T1-w magnetic resonance imaging (MRI). To further evaluate the discriminability of the discovered population-specific profiles, we use the heat-traces of fused brain multigraphs to train and test a linear support vector machine (SVM) classifier using 5-fold cross-validation. This work presents the first step towards `hearing' the shape of the brain multigraph that can be leveraged for profiling and disentangling comorbid neurological disorders, thereby advancing precision medicine.

\begin{figure}[htb!]
\centering
{\includegraphics[width=12.5cm]{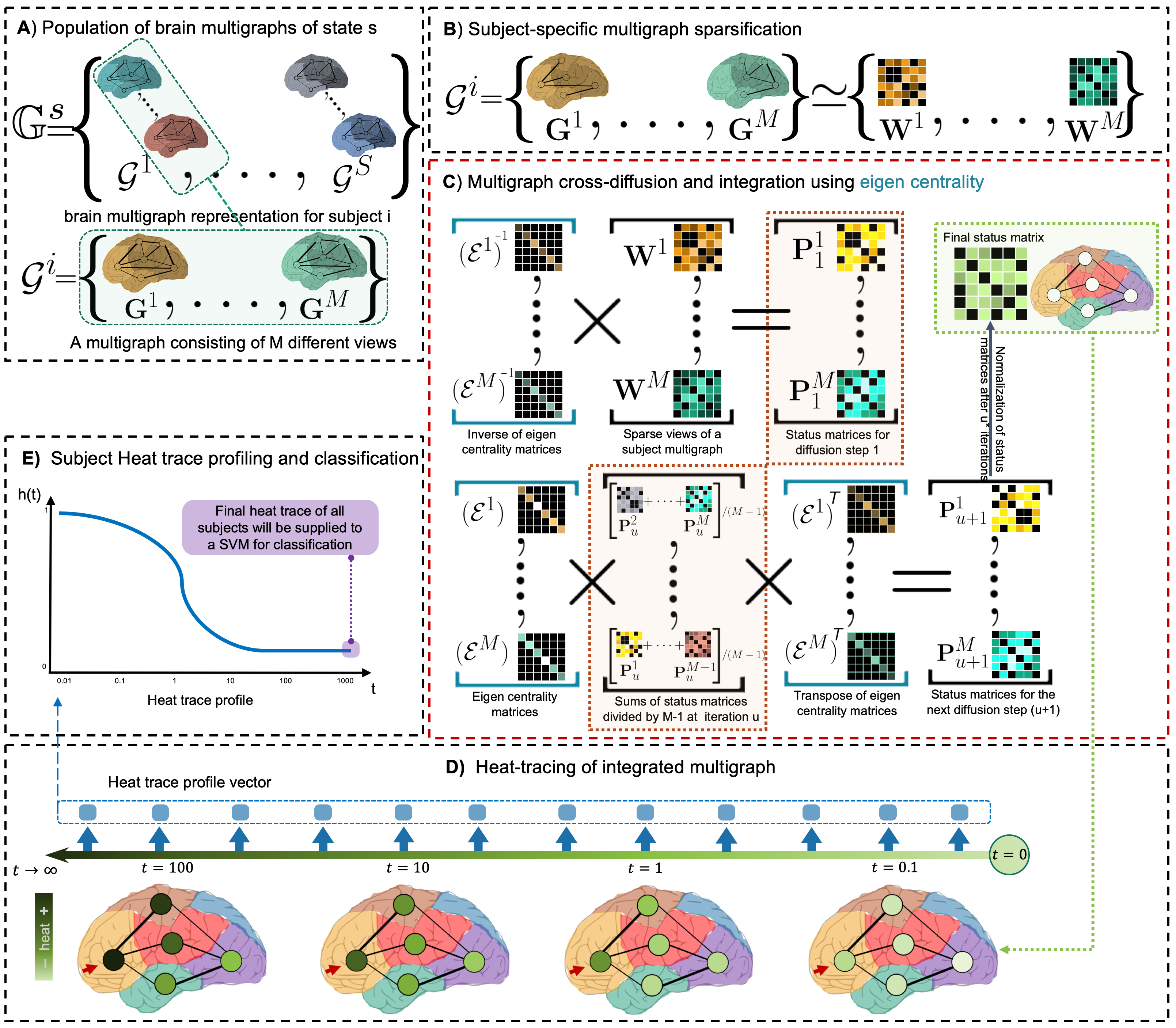}}
\caption{\emph{Proposed eigen-based multigraph cross-diffusion for profiling and comparing brain multigraphs.} \textbf{A)} Dataset $\mathbb{G}^s$ of $S$ brain multigraphs of state $s$ (e.g., disordered or healthy), each represented as a tensor $\mathcal{G}$  with $M$ frontal views encoded in symmetric connectivity matrices $\{ \mathbf{W}^1, \dots, \mathbf{W}^M \}$. \textbf{B)} To remove noisy connectivities and for a more effective graph cross-diffusion, we sparsify each brain graph in $\mathcal{G}$. \textbf{C)} Proposed brain multigraph cross-diffusion and fusion using eigen centrality to produce the integrated multigraph (i.e., status matrix).  \textbf{D)} For each node $v$ in the fused multigraph, we heat the final status matrix using its Laplacian matrix at different timescales $t$. The red arrow points at the active node $v$. \textbf{E)} Heat-trace based profiling and classification. For a given subject $i$, we average the heat traces across all nodes, producing a \emph{time-dependent} heat trace $h(t)$ stored in a heat trace profile vector. By extracting the final heat-traces of all training profiles and supplying them to a support vector machine (SVM), we evaluate the discriminative power of our approach in disentangling different brain states.}
\label{fig:1}
\end{figure}

\section{Proposed Eigen-based Cross-Diffusion and Heat Tracing of Brain Multigraphs}

\textbf{Problem statement.} Given a population $\mathbb{G}^s = \{\mathcal{G}^1,\dots,\mathcal{G}^S \}$ of $S$ brain multigraphs of state $s$, we aim to profile the brain state of the given population $\mathbb{G}^s$ by graph cross-diffusion and Laplacian-based heat tracing. To this aim, we first propose an eigen-based cross-diffusion to integrate each individual brain multigraph into a single graph. Second, we heat the fused graph by Laplacian spectral decomposition and discover the profile of a given population $\mathbb{G}^s$ by averaging all subject-specific heat tracing profiles. In this section, we detail the steps of our eigen-based cross-diffusion for multigraph integration, profiling and comparison framework. In \textbf{Fig}~\ref{fig:1}, we present a flowchart of the five proposed steps including: A) representation of an individual brain multigraph, B) subject-specific sparsification of brain multigraphs, C) cross-diffusion and integration of a multigraph using eigen centrality, D) heat-tracing the integrated multigraph, and E) heat-trace profiling and classification of brain multigraphs.

\textbf{A- Subject-specific brain multigraph representation.} Let $\mathcal{G}^i = \{\mathbf{G}^1,\dots,\mathbf{G}^M \}$ denote a brain multigraph of subject $i$ in the population $\mathbb{G}^s$, composed of $M$ fully-connected brain graphs where $\mathbf{G}^m$ represents the brain graph derived from measurement $m$ (e.g., correlation in neural activity or similarity in morphology). Each brain graph $\mathbf{G}^m \in \mathcal{G}^i$ captures a \emph{connectional view} of the brain wiring. Particularly, we define a brain multigraph $\mathcal{G}^i= (V,\mathcal{W})$ as a set of nodes $V$ representing brain regions of interest (ROIs) across all views and $\mathcal{W} = \{\mathbf{W}^1,\dots,\mathbf{W}^M \}$ is a set of symmetric brain connectivity matrices encoding the pairwise relationship between brain ROIs.  

\textbf{B- Subject-specific sparsification of multigraphs.} Prior to the multigraph diffusion and fusion at the individual level, we first sparsify each brain graph $\mathbf{G}^m$ using different sparsification thresholds for the two following reasons. First, the brain wiring is sparsely inter-connected system where strong connectivity within modules supports specialization whereas sparse links between modules support integration \cite{Rubinov:2011} and weak connectivity weights might not capture well the most important connectional pathways in the brain for the target diffusion task. Hence, we remove the weak connections by sparsifying each brain graph independently. Second, diffusion on fully-connected graphs will rapidly converge to a constant which prohibits a fine-grained characterization of graph topologies to diffuse among one another \cite{Hammond:2013}. Specifically, for every subject $i$ and each view $m$, we vectorize its connectivity matrix $\mathbf{W}^m $ by taking the elements in the off-diagonal upper triangular part. Next, we compute the average mean $\mu_m$ and standard deviation $\sigma_m$ for each view $m$ across all $S$ subjects in $\mathbb{G}^s$. We also define a set of increasing $\alpha$ coefficients, $\alpha = \{\alpha_1, \dots, \alpha_p\}$ to generate $p$ sparsification thresholds $\rho_m^p = \mu_m + \alpha_p \sigma_m$ for each brain graph $\mathbf{G}^m$. Ultimately, for each view, we sparsify all brain graphs. For easy reference, we keep the same mathematical notation $\{\mathbf{W}^1,\dots,\mathbf{W}^M \}$ for the sparsified multigraph adjacency matrices at fixed thresholds $\{ \rho_m^p \}_{m=1}^M$, respectively (\textbf{Fig.}~\ref{fig:1}--B).


\textbf{C- Cross-diffusion and integration of a multigraph using eigen centrality.} Given a \emph{sparsified} brain multigraph $\mathcal{G}^i$ of subject $i$, one can leverage the conventional graph cross-diffusion method introduced in \cite{Wang:2012} to diffuse each brain graph across the average of the remaining brain graphs --progressively altering the individual brain topology in such a way that it resembles more the `average' brain topology. Following the iterative cross-diffusion step, one can integrate all \emph{diffused} graphs by simply linearly averaging them as they lie locally near to one other in the diffused graph manifold. Although compelling, such a technique only relies on the node degree to define the normalized diffusion kernel, which is a limited measure of graph topology that can only capture the local neighborhood of a node in terms of quantify (i.e., number of its neighboring nodes). To better preserve the graph topology during the diffusion process, we unprecedentedly introduce a graph diffusion strategy rotted in eigen centrality, a measure of the influence of a node in a graph based on its eigen centrality. An eigen central node is directly related to nodes which are central themselves \cite{Grassi:2007}. Hence, it presents a stronger definition of graph centrality taking into account the entire pattern in a graph, which is also an intrinsic property of brain networks \cite{Joyce:2010}. Eigen centrality is a function of the connections of the nodes in one's neighborhood \cite{Bonacich:1972}. For a single view $m$, let $\{ \lambda_1, \dots, \lambda_{|V|} \}$ denote the set of eigenvalues of the graph adjacency matrix $\mathbf{W}^m$ and $\psi = max_i |\lambda_i|$ its spectral radius, the eigen centrality of the $k^{th}$ node in $\mathbf{G}^m$ is defined as the $l^{th}$ entry component of the principal eigen vector $\mathbf{x}$, that is $\mathbf{x}_l = \frac{1}{\psi} \sum_{k=1}^{|V|} \mathbf{W}^m(l,k) \mathbf{x}_k$ \cite{Bonacich:1972}, where $|V|$ denotes the number of nodes in the graph. Next, for each graph in $\mathcal{G}^i$, we define a diagonal matrix $\mathcal{E}^m$ for its $m^{th}$ view storing graph node eigen centralities. This will be used to define an \emph{eigen centrality-normalized} connectivity matrix $\mathbf{P}^m$ as follows (\textbf{Fig.}~\ref{fig:1}--C): $\mathbf{P}^m = \mathcal{E}^{m^{-1}} \mathbf{W}^m$. 

Next, for each view $m$, we iteratively update the status matrix $\mathbf{P}^m$ through diffusing the average global structure of other $(M-1)$ views of the brain multigraph $\mathcal{G}^i$ along the eigen centrality diagonal matrix $\mathcal{E}^m$, thereby forcing the connectivity diffusion to go through the most central ROIs in the brain. As such, we cast a new formalization of \emph{edge-based} diffusion on graphs guided by most central nodes, which may overlook noise that distributes randomly and sparsely in $\mathbf{G}^m$ as well as irrelevant connections. At iteration $u+1$, we use the following update rule to compute the status matrix of $\mathbf{G}^m$:

\begin{gather}
    \mathbf{P}_{u+1}^m = \mathcal{E}^m \times (\frac{1}{M-1}\sum_{k \neq m}\mathbf{P}_u^k) \times (\mathcal{E}^m)^T
\end{gather}

Following $u^\ast$ iterations of graph cross-diffusion, we then produce the fused brain multigraph $\mathbf{P}_{u^\ast}^i$ for subject $i$ by linearly averaging the view-specific status matrices as follows: $\mathbf{P}_{u^\ast}^i = \frac{1}{M}\sum_{m=1}^M \mathbf{P}^m_{u^\ast}$. 


\textbf{D- Subject-specific heat-tracing of the fused multigraph.} In this stage, given the fused status matrix $\mathbf{P}_{u^\ast}^i$, we set out to define a continuous time-dependent profile (i.e., curve) of the fused brain multigraph of subject $i$ using a \emph{node-based} diffusion process. Inspired from the work of \cite{Kac:1966}, we leverage the graph spectrum encoding its shape (or geometry) to profile $\mathbf{P}_{u^\ast}^i$. To this aim, we first define the normalized Laplacian matrix $\mathcal{L}^i$ of the final status matrix $\mathbf{P}_{u^\ast}^i$ as $\mathcal{L}^i = \mathbf{I}- \mathbf{S}^\frac{-1}{2}  \mathbf{P}_{u^\ast}^i \mathbf{S}^\frac{-1}{2}$, where $\mathbf{S}$ is the diagonal strength matrix and $\mathbf{I}$ is the identity matrix. Second, we estimate the spectrum of the normalized Laplacian $\mathcal{L}^i$ with eigenvalues $\{\lambda_1^i,\dots,\lambda_{|V|}^i\} $ (\textbf{Fig.}~\ref{fig:1}--D). Next, we use the heat equation to heat a node $v$ in the fused multigraph at timescale $t$ as follows: $h_t(v) = \sum_{k=1}^{|V|}{e^{-t\lambda_k}}(v)$, which is also referred to as the \emph{heat trace} of node $v$ \cite{Tsitsulin:2018}. By averaging the \emph{heat traces} of all nodes in the fused brain multigraph, we can estimate its heat trace at time $t$. Ultimately, we create a logarithmic sample space spanning from $10^{-2}$ to $10^{3}$ to better inspect the descend of heat-traces by acceleratingly increasing timescales. For $n_t$ different timescales in the logarithmic space, we compute $n_t$ different averaged heat-traces to create a heat trace vector $[h^i_{t^1}, h^i_{t^2}, \dots, h^i_{t^{n_t}}]$ and profile the fused multigraph of a subject $i$. The steps of our method are detailed in \textbf{Algorithm}~\ref{algo:1}.

\begin{algorithm}
\scriptsize
\caption{Eigen-based cross-diffusion for multigraph integration and profiling}
\begin{algorithmic}[1]
    \State{\textbf{INPUTS}:\newline \indent $\mathcal{G}^i= \{\mathbf{G}^1,\dots,\mathbf{G}^M\}$: multigraph of the $i^{th}$ subject in dataset $\mathbb{G}^s$ of state $s$ }
    \For{$m := 1$ \textbf{to} $M$}
        \State{$\mathbf{W}^m \gets$ matrix representation of graph $\mathbf{G}^m$ ($m^{th}$ view of $\mathcal{G}^i$})
        \State{\textcolor{blue}{$\mathcal{E}^m$} $\gets$ diagonal matrix built from eigen centralities of $\mathbf{W}^m$}
        
        \State{$\mathbf{P}^m_1 \gets$ \textcolor{blue}{$\mathcal{E}^{m^{-1}}$} $\mathbf{W}^m$} (\textcolor{blue}{eigen centrality normalization}) \Comment{first status matrix}
        \EndFor
    \For{each diffusion iteration $u \in \{1,2,\dots,u^\ast\}$}
        \For{$m := 1$ \textbf{to} $M$}
            \State{Update the status matrix of the $m^{th}$ view via cross-diffusion using \newline 
            \indent\indent\indent $\mathbf{P}_{u+1}^m \gets$ \textcolor{blue}{$\mathcal{E}^m$} $\times (\frac{1}{M-1}\sum_{k \neq m}\mathbf{P}_u^k) \times$ \textcolor{blue}{$(\mathcal{E}^m)^T$}}
        \EndFor
    \EndFor
    \State{Compute the final status matrix for subject $i$ using $\mathbf{P}_{u^\ast}^i \gets \frac{1}{M}\sum_{m=1}^M \mathbf{P}_{u^\ast}^{m}$}
    \State{$\mathcal{L}^i \gets $ normalized Laplacian matrix of $\mathbf{P}_{u^\ast}^i$ of subject $i$}
    \State{$\{\lambda_1^i, \dots, \lambda_k^i\} \gets$ eigenvalues of Laplacian $\mathcal{L}^i$ of subject $i$}
    \For{each logarithmic timescale $t \in \{t^1,t^2,\dots,t^{n_t}\}$}
        \For{each node $v$ of the normalized Laplacian $\mathcal{L}^i$ of subject $i$}
            \State{Compute heat-trace $h_t(v) = \sum_{k=1}^{|V|}{e^{-t\lambda_k^i}}(v)$}
        \EndFor
        \State{Compute time-dependent heat-trace $h_t^i$ for subject $i$ averaged across subject nodes at timescales $t$}
    \EndFor
     \State{\textbf{OUTPUTS}: heat-trace vector of subject $i$, $[h^i_{t^1}, h^i_{t^2}, \dots, h^i_{t^{n_t}}]$}
\end{algorithmic}
\label{algo:1}
\end{algorithm}

\textbf{E- Heat-trace profiling and discriminability of brain profiles.} Given a population $\mathbb{G}^{s_1}$ of brain multigraphs of state $s_1$ (e.g., healthy) and a population $\mathbb{G}^{s_2}$ of state $s_2$ (e.g., disordered), we compute the heat trace profile for each brain multigraph in each population. Next, we report the average population heat trace profile by averaging the profiles of all individual multigraphs in the population. To evaluate the discriminability of the estimated fused multigraph heat tracing profiles, we train a support vector machine (SVM) with a sigmoid kernel classifier to classify brain multigraphs in state $s_1$ or $s_2$ using the stable heat trace value at the tail of the profile curve (\textbf{Fig.}~\ref{fig:1}--E). Specifically, we use 5-fold cross-validation to train an SVM classifier using the single-valued heat trace of each subject $i$. We also define a margin $\delta(s_1,s_2)$ between the two brain states by computing the absolute difference between the heat trace value at the tail of both heating profiles.

\begin{figure}[htb!]
\centering
{\includegraphics[width=12.5cm]{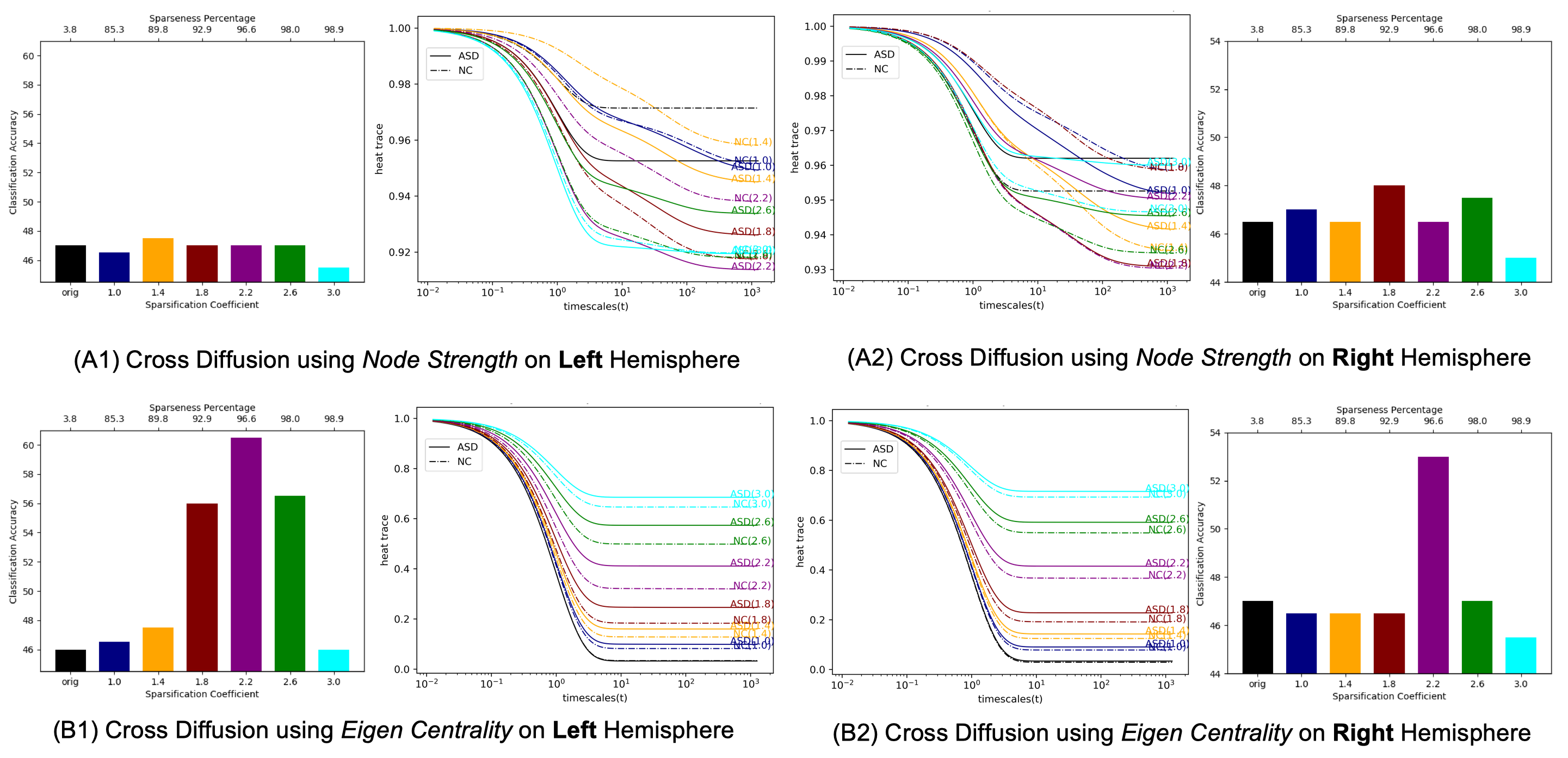}}
\caption{\emph{Heat-trace profiles of morphological brain multigraphs with healthy and autistic states and SVM classification results using both proposed eigen-based cross-diffusion and conventional strength based cross-diffusion method \cite{Wang:2012}.} NC: normal controls. ASD: autism spectrum disorders. Clearly, our method produces orderly and smooth heat-trace profiles with larger gaps between brain states, whereas the conventional method produces fluctuating and wavy profiles. This nicely results in our method achieving higher classification accuracy at different sparsification thresholds, thereby demonstrating the discriminativeness of the estimated profiles.}
\label{fig:2}
\end{figure}

\section{Results and Discussion}

\textbf{Brain multigraph dataset and parameter setting.} We evaluated our framework on 200 subjects (100 ASD and 100 NC) from Autism Brain Imaging Data Exchange (ABIDE). For each cortical hemisphere, each subject is represented by 4 cortical morphological brain networks derived from maximum principal curvature, the mean cortical thickness, the mean sulcal depth, and the average curvature. These networks were derived from T1-weighted magnetic resonance imaging (MRI). Each hemisphere was parcelled into 35 anatomical regions defining the nodes of each brain graph and encoded in a symmetric matrix that quantifies morphological dissimilarity between pairs of cortical regions using a particular measurement (e.g., cortical thickness) \cite{Mahjoub:2018,Soussia:2018b,Nebli:2020}. Hence, each cortical hemisphere is represented by a multigraph consisting of 4 different graphs. 

\emph{Brain multigraph sparsification.} For each hemisphere, we set the sparsification coefficients $\alpha$ to $\{1.0, 1.4, 1.8, 2.2, 2.6, 3.0\}$ to sparsify the 4 brain graphs in each multigraph. Next, we plot the average heat-trace profile across subjects in the same population (i.e., sharing the same state) and report SVM classification results using 5-fold cross-validation in \textbf{Fig.}~\ref{fig:2}.

\textbf{Evaluation and comparison methods.} We compare the performance of our eigen-based cross-diffusion framework with conventional strength based cross-diffusion method \cite{Wang:2012}. As conventional cross-diffusion method uses a diagonal matrix storing node strengths on the diagonal, it cannot capture the quality of the local neighborhoods (e.g., presence of hub neighbors); whereas our method is based on eigen centrality measures which assesses the quality of local neighbors to a given node. \textbf{Fig.}~\ref{fig:2} shows that \cite{Wang:2012}  produces unstable and highly fluctuating heat-trace plots, whereas our eigen-based cross-diffusion method generates ordered and smooth heat-trace plots for ASD and NC brain populations $\mathbb{G}^{{ASD}}$ and $\mathbb{G}^{{NC}}$. This can be explained by the fact that \cite{Wang:2012} diffuses a sparse similarity matrix encoding node similarity to nearby data points, whereas we diffuse the eigen diagonal matrix which enhances the role of hub nodes as reliable mediators of information diffusion which cannot be captured by only considering nearest neighbors. Besides, central nodes are generally more resistant to noise which can permeate local neighborhoods, thereby privileging their use for stable and robust diffusion. The orderliness of our method shows its true power in brain state classification by SVM. Our proposed method of eigen-based cross-diffusion boosts the classification results by 2-12\% in comparison with baseline method.   

\begin{figure}[htb!]
\centering
{\includegraphics[width=12cm]{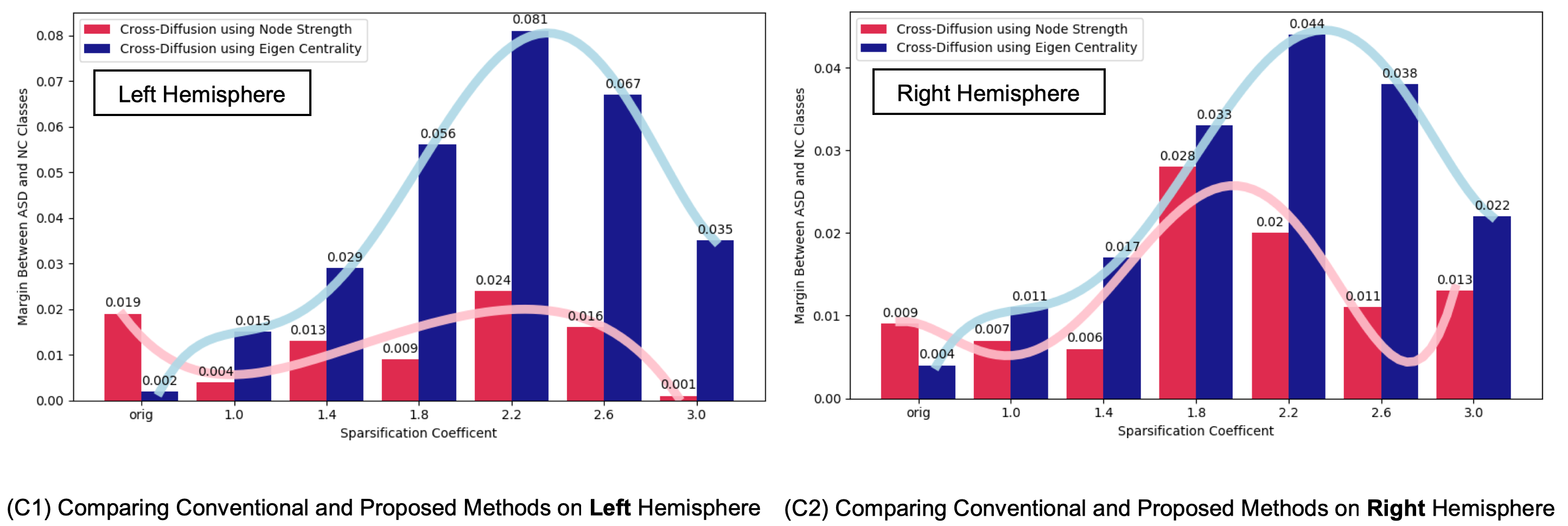}}
\caption{\emph{Comparison the margin $\delta(ASD, NC)$ between ASD and NC classes shown in \textbf{Fig.}~\ref{fig:2} at different sparsification levels by our method and \cite{Wang:2012}}. We fitted $5^{th}$ degree polynomials to the bar plots.}
\label{fig:3}
\end{figure}

In \textbf{Fig.}~\ref{fig:3}, we display the margin $\delta(ASD, NC)$ between two brain populations for right and left hemispheres at different sparsification thresholds. Clearly, our method produces larger gaps between autistic and healthy brain state profiles (blue bars), which demonstrates its discriminative potential for neurological disorder diagnosis and classifying brain states. As we increase the sparsification level of brain multigraphs, the gap first increases then decreases fitting a smooth polynomial curve. We also note that the margin is very low $\delta(ASD, NC)$ when using the original non-sparsified brain multigraphs, implying lower state discriminativeness. In fact, the sparsification threshold is a hyper-parameter that requires a deeper investigation. Ideally, one would learn how to identify the best threshold that allows to identify what individualizes a population of brain multigraphs.

\section{Conclusion}

In this work, we introduced a multigraph cross-diffusion, integration and profiling technique based on eigen centrality. The discovered brain multigraph profiles were smooth and highly discriminative in comparison with baseline method, which have utility in diagnosing neurological disorders. Indeed, the wide spectrum of the disordered brain connectome \cite{Fornito:2015} demands not only advanced graph analysis techniques and scalable graph comparison strategies, but it also calls for new multigraph analysis tools that can unify the multiple graph representations of the brain including structure and function. In our future work, our goal is to profile a wide spectrum of brain disorders using functional, structural and morphological brain graphs in future population comparative connectomics \cite{Heuvel:2016,arbabshirani2013functional}.

\section{Supplementary material}

We provide a supplementary item for reproducible and open science:

\begin{enumerate}
	\item An 8-mn YouTube video explaining how our multi-graph profiling framework works on BASIRA YouTube channel at \url{https://youtu.be/D_E2m6O37mk}. 
\end{enumerate}

\section{Acknowledgments}

I. Rekik is supported by the European Union's Horizon 2020 research and innovation programme under the Marie Sklodowska-Curie Individual Fellowship grant agreement No 101003403 (\url{http://basira-lab.com/normnets/}).

\bibliography{Biblio3}
\bibliographystyle{splncs}
\end{document}